\documentclass{article}
\usepackage{ijcai19}

\usepackage{times}
\usepackage{soul}
\usepackage{url}
\usepackage[hidelinks]{hyperref}
\usepackage[utf8]{inputenc}
\usepackage[small]{caption}
\usepackage{graphicx}
\usepackage{amsmath}
\usepackage{booktabs}
\usepackage{algorithm}
\usepackage{algorithmic}
\usepackage{color}
\usepackage{amsmath}
\usepackage{setspace}

\usepackage{tabularx}
\usepackage{color}
\usepackage{array}
\newcolumntype{L}[1]{>{\raggedright\let\newline\\\arraybackslash\hspace{0pt}}m{#1}}
\newcolumntype{C}[1]{>{\centering\let\newline\\\arraybackslash\hspace{0pt}}m{#1}}
\newcolumntype{R}[1]{>{\raggedleft\let\newline\\\arraybackslash\hspace{0pt}}m{#1}}
\title{Energy-Efficient Slithering Gait Exploration for a Snake-like Robot \\
	based on Reinforcement Learning}

\author{
Zhenshan Bing$^1$
\and
Christian Lemke$^2$\and
Zhuangyi Jiang$^1$\and
Kai Huang$^{3}$\And
Alois Knoll$^1$
\affiliations
$^1$Department of Computer Science, Technical University of Munich, Germany\\
$^2$Department of Computer Science, Ludwig Maximilian University of Munich, Germany\\
$^3$School of Data and Computer Science, Sun Yat-sen University, China\\
\emails
bing@in.tum.de,~
Christian.Lemke@campus.lmu.de,~
jiangz@in.tum.de,\\
huangk36@mail.sysu.edu.cn,~
knoll@in.tum.de
}

\hypersetup{draft}

\begin{document}
	
\maketitle
%\\[-13.0ex]

\begin{abstract}
Similar to their counterparts in nature, the flexible bodies of snake-like robots enhance their movement capability and adaptability in diverse environments. 
However, this flexibility corresponds to a complex control task involving highly redundant degrees of freedom, where traditional model-based methods usually fail to propel the robots energy-efficiently.
In this work, we present a novel approach for designing an energy-efficient slithering gait for a snake-like robot using a model-free reinforcement learning (RL) algorithm.
Specifically, 
we present an RL-based controller for generating locomotion gaits at a wide range of velocities, which is trained using the proximal policy optimization (PPO) algorithm.
Meanwhile, a traditional parameterized gait controller is presented and the parameter sets are optimized using the grid search and Bayesian optimization algorithms for the purposes of reasonable comparisons.
Based on the analysis of the simulation results, we demonstrate that this RL-based controller exhibits very natural and adaptive movements, which are also substantially more energy-efficient than the gaits generated by the parameterized controller.
Videos are shown at \textcolor{blue}{\href{https://videoviewsite.wixsite.com/rlsnake}{https://videoviewsite.wixsite.com/rlsnake}}.

\end{abstract}

\section{Introduction}

Snake-like robots, as a class of hyper-redundant mechanisms, carry the potential of being one kind of promising mobile robotic applications that are capable of traveling and performing tasks in diverse environments, such as disaster rescue, underwater exploration, and industrial inspection~\cite{liljeback2012snake}.
Since snake-like robots can only carry limited energy resources for field operations, it is important to develop energy-efficient gaits to reduce the impact of the power constrains.
On one hand, optimizing the power consumption can prolong the service time of a robot and maximize its locomotion performance at the same time.
On the other hand, a sufficient energy system may in return allow us to design a more lightweight robot or add other functional components~\cite{tesch2009parameterized}.
However, it is challenging to design energy-efficient gaits for snake-like robots on the basis of their redundant degrees of freedom (DOF) and the complex interactions with the environment~\cite{Tucker1975energetic}.

Since the first snake-like robot was built in $1972$, researchers have been working constantly on designing more advanced snake-like robots~\cite{liljeback2012snake} and sophisticated gaits for robots with different types of mechanical configurations or terrains.
Meanwhile, slithering gait has been considered as the most promising gait for snake-like robots to perform autonomous locomotion tasks, which imitates the serpentine locomotion of real snakes~\cite{hu2009mechanics}.
Hirose first used the \textit{serpenoid curve} to control a snake-like robot, which was an effective approach by imitating the real snake movement~\cite{hirose1993biologically}.
Ma proposed another model \textit{serpentine curve} to describe the locomotion of snakes by modeling their muscle characteristics and achieved a higher locomotive efficiency than the \textit{serpenoid curve} by running simulations~\cite{774054}.
On the basis of these snake-like movement curves, the \textit{gait equation}, as a robust and effective method, works as an abstract expression of gaits of a snake-like robot by describing joint angles as parameterized sinusoidal functions~\cite{tesch2009parameterized}. 
It allows for the emergence of complex behaviors from low-dimensional representations with
only few key parameters, greatly expanding their maneuverability and simplifying user control.
With this method, researchers developed several biological gaits for snake-like robots to move in the indoor and outdoor environment~\cite{7017664}.

However, optimizing these parameterized gaits for the purpose of energy saving is difficult and limited, since they are confined to those abstracted gait parameters and only few studies have been reported.

Crespi et al. adopted a heuristic optimization algorithm to rapidly adjust the travel speed of the robot~\cite{4459741}.
Tesch et al. used the Bayesian optimization approach to regulate those open-loop gait parameters for snake robots, which made the robot move faster and sturdier~\cite{6095076}.
Gong et al. proposed a shape basis optimization algorithm to simplify the gait design parameter space and came up with a novel gait in granular materials~\cite{gong2016simplifying}.
Even so, all these works still focus on optimizing the gait on the basis of the parameterized gait generation system, and have very limited effect on further improving gait efficiency.

This gait optimization task, however, corresponds to a complex control problem due to two primary reasons~\cite{liljeback2012snake}.
The extrinsic challenge comes from the complex dynamic interaction between the ground and the redundant mechanism with many degrees of freedom.
Therefore, it is extremely important to model precisely and rapidly.
The intrinsic challenge is how to synchronize and coordinate all the body joints to exhibit a proper motion pattern integrally, which is expected to be both robust and efficient.

As an emerging technology, reinforcement learning (RL) reveals the nature of the learning process of locomotion in animals that can offer a model-free learning to master new skills or adapt to diverse environments.
On this basis, this work offers a novel alternative to design the slithering gait of a snake-like robot based on reinforcement learning technology.
Our main contributions are summarized as follows.
First, we define the energy-efficiency metrics and introduce the parameterized slithering gait design method as the baseline, which is optimized by using a grid search method and the Bayesian optimization method in terms of energy efficiency.
Second, we propose a gait controller using the state-of-the-art RL algorithm PPO in simulation. 
The learned gait exhibits surprising similarity to the natural movement of real snakes.
Third, the results demonstrate that the learned gait successfully outperforms the parametrized slithering gait in terms of energy efficiency at a range of velocities.

\section{Related Work}
\textcolor{black}{
	Unlike other land animals, snakes achieve diverse locomotion gaits by twisting their bodies on various terrain and exhibit undulatory locomotion.
	To imitate similar efficient movement, most snake-like robots are controlled by the kinematics-based method, which can be regarded as a process to simplify parametric representations of the snake-like trajectories~\cite{hirose1993biologically,774054,tesch2009parameterized}.
	However, this method limits gait efficiency by only manually tuning those parameters, even without considering that it is time-consuming and inefficient.  
}

\textcolor{black}{
	Although there are few studies on optimizing the gaits for snake-like robots, studies on optimizing the locomotion gaits of other kinds of robots have been reported for the purpose of energy saving.
	Initially, researchers adapted techniques for multidimensional function optimization tasks to design efficient gaits, such as evolutionary algorithms~\cite{chernova2004evolutionary} or policy gradient search algorithms~\cite{kim2003automatic,kohl2004machine}.
	However, these algorithms are usually plagued by local optima, which makes the process slow, inefficient, and manually intensive.
	Afterwards, gait optimization methods based on prior knowledges are further investigated.
	Lizotte et al. first optimized the speed and smoothness of the gait with Gaussian process regression on a quadruped robot\cite{lizotte2007automatic}. 
	Calandra et al. used Bayesian optimization approach to regulate those open-loop gait parameters for bipedal robots, which made the robot move faster and robuster~\cite{calandra2014bayesian}.
	Even though these optimized gaits outperform the hand-coded gaits, they are still very inefficient forms of locomotion compared with the natural movement achieved by animals.
}

As an intelligent trial-and-error learning method, RL brings new solutions for free gait generation tasks without knowing precise models or prior knowledge.
Initially, RL was not widely used in the domain of robotics often presented with high-dimensional, continuous states and actions~\cite{kober2013international}.
However, with more and more advanced RL algorithms coming out, many robotic implementations are able to handle complicated tasks, such as gait generations~\cite{peng2017deeploco}, dexterous manipulation~\cite{rajeswaran2017learning}, and autonomous driving~\cite{8461113}.  
In \cite{cully2015robots}, two prototype experiments are shown in which RL can help robots recover from damage and adapt quickly like animals do.
A hexapod robot learns to walk fast and straight with broken or missing legs.
And a robotic arm learns to reach its previous position goal with one or more stuck joints.
Afterwards, RL technologies are increasingly used in free gait generation tasks.
Schulman et al.~\cite{schulman2017proximal} implemented their PPO algorithm on a collection of benchmark robotic locomotion tasks from 2D swimmer to 3D humanoid robot.
Except for simply using RL-based methods to generate gaits for robots, they have been used to learning energy-efficient gaits.  
Kormushev et al.~\cite{Kormushev2018} used RL to optimize the vertical center-of-mass trajectory and the walking pattern of a bipedal robot to exploit the passive compliance for the purpose of energy efficient walking. 
Yu et al.~\cite{yu2018learning} proposed an RL-based approach for creating low-energy and speed-appropriate locomotion gaits for four types of walking, including biped walking, quadruped galloping, hexapod walking, and humanoid running.

In this paper, we focus on using RL to learn slithering gaits for a snake-like robot with redundant degrees of freedom, so that the learned gaits can outperform those parameterized gaits in terms of energy efficiency.

\section{Models and Metric Definition}

In this section, we first introduce the snake-like robot model used for exploring different gaits.
Then, we present our energy efficiency metric for comparing different gaits based on our robot model.

\subsection{Snake-like Robot Model}

\begin{figure}[tbp]
	\centerline{
		\includegraphics[width=0.42\textwidth]{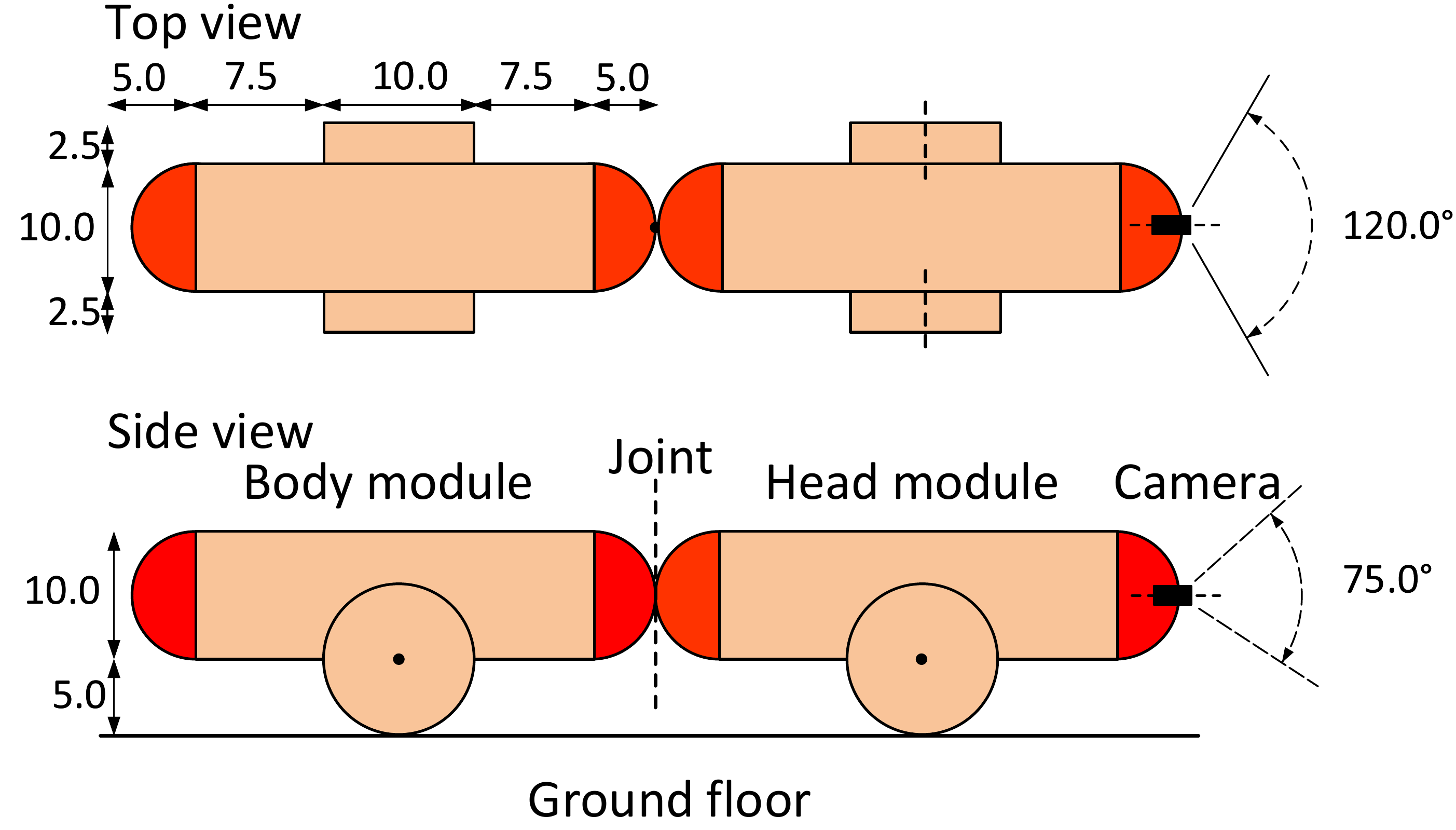}}
	\caption{The top view and side view of the first two modules of the snake-like model (in centimetre).}
	\label{fig:snake_model_top}
\end{figure}

We model and simulate our snake-like robot in MuJoCo~\cite{6386109}, which is a physics engine offering fast and accurate robot simulation environment.
The snake-like robot model used in this study is inspired by the \textit{ACM} snake-like robot~\cite{hirose1993biologically}, which uses eight joints and nine identical modules.
The first module is used as the head module and equipped with a vision sensor for performing other tasks.
Figure~\ref{fig:snake_model_top} shows the technical drawings of the first two modules of the snake-like robot model.  
The model acts in a dynamic environment, therefore its weight, friction on the ground, and actuator power are critical to enable availability. 
A uniform density of $600~kg/m^3$ is set for all components of the model. 
This density is selected based on the robot developed by Dowling~\cite{dowling1996limbless}, which has about the same value including all mechanical and electrical components. 

\subsubsection{Actuated Joints}
The snake-like robot adopts actuated joints to bend its body to slither forward. 
All the joints are modeled as a servo motor and rotate along the perpendicular direction to the ground in a range of $[-90^{\circ}, 90^{\circ}]$.
The force is limited to a range of $[-20, 20]$ in Newton, which is adequate to propel the modules at a reasonable speed and strength. 
The maximum actuator torque can be calculated by multiplying the actuator force with the gear length of $0.175~m$, which is the half length of a module.

It is worth mentioning that a configuration with torque motor is also tested and works as well as the servo motors. 
For comparison purposes, the servo motor type is chosen because the \textit{gait equation} controller also depends on this input type, which will be further explained in Section~\ref{sec_param}.

\subsubsection{Passive Damping Wheels}
Each robot module is equipped with two passive wheels, which are used to imitate the anisotropic friction property of the snake skin. 
Like the movement of real snakes, those passive wheels enable a minimum friction in the direction of rotation and a high friction in the lateral direction.
%how to create Anisotropic friction
% Is it called damping ???? damping is something for springs?
% TODO maybe by how much
%Real snakes do not roll or slide in a forward direction while moving. 
%However, such movement would be possible with free rolling wheels. 
%In fact, such locomotion behavior had been observed. 
%Without this limitation, the RL algorithm tends to slither to accelerate, before forming a straight line with its body and using the rolling effect. 
%However, the robot tends to form itself in a line and rolls forward.
To enable the torque energy costs at all the joints, a constant damping property is added at each wheel joint. 
This constantly decelerates the rotation of the wheels, which imitates a low friction in forward direction. 
Otherwise, the robot will just form itself in a straight line and roll forward to gain distance while avoiding joint torque energy costs. 

%In MuJoCo, this decelerating effect is set by the damping property of the wheels. 

\subsection{Energy Efficiency Metric}
Mobile robots have to conserve their battery power so they can operate for long periods of time. 
Our goal is to design gaits that are capable to move in an energetic and economical manner and are still versatile to move at a range of velocities~\cite{dowling1997power}.

\subsubsection{Power Profiles}
The power profile is one set of the average power for each module of the robot during a test run, which offers detailed information to help improve the gait generations. 
For example, we can observe that which modules consume more power and therefore are more likely to heat up fast or broken. 

For a snake-like robot with $N$ joints, the averaged power $\bar{P}_j$ for the $j^{th}$ actuator is the averaged absolute value of the product of the torque $\tau_j$ and its angular velocity
$\dot{\phi}_j$ during a run with $k$ steps.
Thus, our first metric can be expressed as
\begin{equation}
\bar{P}_j = \frac{1}{k} \sum_{1}^{k} \left | \tau_j \dot{\phi}_j  \right |, \forall j \in[1, N] \text{.}
\end{equation}
%
%The first and most straightforward measure of power usage for the $j^{th}$ actuator $\epsilon_j$ is the absolute value of the product of the torque $\tau_j$ and its angular velocity
%$\dot{\phi}_j$. 
\subsubsection{Power Efficiency}
Based on the power consumption of each module of the robot, the total power consumption $P$ of all $N$ actuators at each time step is calculated by
\begin{equation}
\centering
P = \sum_{j=1}^{N} \left | \tau_j \dot{\phi}_j  \right | \text{,}
\end{equation}
where the torque $\tau_j$ is the product of its applied force $f_j$ and its gear constant parameter $h_j$ (the length of the actuator). 
The model uses actuators with a limited force of $f_{max}$ as maximum force in both directions.
With this property, the normalized power consumption $\hat{P}$ is calculated as
\begin{equation}
\centering
\hat{P} = \frac{1}{N} \sum_{j=1}^{N}{\frac{\left | f_j h_j \dot{\phi}_j \right | }{f_{max} h_j \dot{\phi}_{max}}}~\text{.}
\label{equ_normal_power}
\end{equation}
This normalized power consumption $\hat{P}$ will be further used for reward definition.

With the variables of energy consumption $P$ and the velocity $v$, several efficiency metrics can be calculated. 
A usual way is to calculate the Cost of Transport (COT), which is the power $P$ or energy $E$ divided by the mass $m$, the gravity $g$ and the distance $d$ or velocity $v$:
\begin{equation}
\centering
COT = \frac{E}{mgd} = \frac{P}{mgv}
\end{equation}

This unit-less measure is also able to compare the efficiency between different mobile systems. 
Since the same robot model and the same environmental properties are compared, those constants only scale the results and would not provide any additional insight for the comparison. 
Thus, the second metrics is to calculate the Averaged Power per Velocity (APPV) as 
\begin{equation}
APPV= \frac{{P}}{v}
\end{equation}
Similar power efficiency metrics can be found in \cite{tesch2009parameterized} and \cite{saito2002modeling}.

\section{Baseline Examples}
\label{sec_param}
This section provides two baseline examples, where the parameterized \textit{gait equation} controller is presented to generate the slithering gait for our snake-like robot.
By searching a grid of gait parameters with fixed intervals, we try to find out the best energy-efficiency gaits at different velocities that can be acquired by this controller.
Then we use the Bayesian optimization algorithm to explore better parameter combinations in the range of searching grid, since the searching grid is relatively sparse. 

\subsection{Gait Equation Controller}
%\begin{table*}[t]
%	\caption{The grid search parameters for the gait generation with the equation controller}
%	\centering
%	\def\arraystretch{1.15}
%	\begin{tabular}{l l r}
%		\toprule 
%		Parameters & Descriptions & Values\\
%		\hline
%		$\omega$ & Angular frequency & 0.25, 0.5, 0.75, 1.0, 1.25, 1.5,\\ 
%		& &  1.75, 2.0, 2.25, 2.5, 2.75, 3.0\\
%		\hline  
%		y (x=1-y) & Linear reduction & 0.1, 0.2, 0.3, 0.4\\
%		\hline  
%		$A$ & Amplitude (in degrees) & 40, 50, 60, 70, 80, 90, 100, 110,\\
%		& & 120, 130, 140, 150, 160, 170, 180\\
%		\hline
%		$\lambda$ & Phase difference (in degrees) & 40, 50, 60, 70, 80,  \\
%		& & 90, 100, 110, 120 \\% Spatial frequency
%		\bottomrule 
%	\end{tabular}
%	\label{tab:grid_search_parameter}
%\end{table*}

\begin{table}[b]
	\centering
	\def\arraystretch{1.1}
	\begin{tabular}{L{1.2cm} L{2.5cm} L{3.5cm}}
		\toprule 
		Params. & Descriptions & Values\\
		\hline
		$\omega$ & Temporal frequency & 0.25, 0.5, 0.75, 1.0, 1.25, 1.5,
		1.75, 2.0, 2.25, 2.5, 2.75, 3.0\\
		\hline  
		y  & Linear reduction & 0.1, 0.2, 0.3, 0.4\\
		\hline  
		x  & Linear reduction & $(1-y)$\\
		\hline  
		$A$ & Amplitude (in degrees) & 40, 50, 60, 70, 80, 90, 100, 110,
		, 120, 130, 140, 150, 160, 170, 180\\
		\hline
		$\lambda$ & Spatial frequency (in degrees)  & 40, 50, 60, 70, 80,
		90, 100, 110, 120 \\% Spatial frequency
		\bottomrule 
	\end{tabular}
	\caption{The gait parameters used for the grid search algorithm.	\label{tab:grid_search_parameter}}
\end{table}

The \textit{gait equation} method represents a kinematic locomotion controller that uses a mathematical equation to describe its gait.
%This method was first proposed as serpentine curve by Hirose who gained inspiration from real snakes.
In this work, an undulation gait equation extended from \cite{tesch2009parameterized} is used for the purpose of comparison. 
The \textit{gait equation} controller is modeled as
\begin{equation}
\phi(n, t) = (\frac{n}{N} x + y) \times A \times \sin (\omega t + \lambda n)~\text{.}
\label{equ_equcontroller}
\end{equation} 
$\phi(n, t)$ presents the joint angle value at time $t$, where $n$ is the joint index and $N$ is the joint amount.
$\lambda$ and $\omega$ are the spatial
and temporal frequency of the movement, respectively. 
The spatial frequency represents the cycle numbers of the wave and
the temporal frequency represents the traveling speed of the
wave.
$A$ is the serpentine amplitude and $x$ and $y$ are the constants for shaping the body curve.

To ensure a fair comparison, we use the grid search method to generate a variety of gaits and find out those parameter combinations with the best power efficiency at different velocities. 
The grid search method generates a Cartesian product from the parameters in Table~\ref{tab:grid_search_parameter}, resulting in $6480$ parameter sets.
Then, each motion parameter set gets tested by running $1000$ steps in the simulation environment.
For each run, the first $200$ time steps are ignored and the remaining $800$ time steps are evaluated for collecting experiment data.
This is done because it has been observed that the snake robot needs about $200$ time steps to accelerate and then moves at a steady speed.

\subsection{Bayesian Optimization Method}
Since the searching grid is relative sparse, we further use the Bayesian optimization algorithm to search for better parameters that may be located in those grid intervals.
This strategy has been used for optimizing snake-like robot gait parameters in~\cite{6095076} and other kinds of robots~\cite{lizotte2007automatic,calandra2014bayesian}.

The purpose of this method is to find a group of optimized parameters in \eqref{equ_equcontroller} to achieve minimum power consumption at different velocities.
The boundaries of these parameters are set as same as the value boundaries in Table~\ref{tab:grid_search_parameter}.
According to the \textit{gait equation} controller, the maximum speed of the robot is mainly bounded by its temporal frequency $\omega$.
Therefore, we perform the optimization process twelve times when $\omega$ is uniformly sampled from $[0,~3.0]$ at a step size of $0.25$, which results in $12$ trials.
For each search, the exploration and exploitation samples are set as $10$ and $100$ for avoiding the local minimum.

\subsection{Baseline Performance}
\begin{figure}[t]
	\centerline{
		\includegraphics[width=0.425\textwidth]{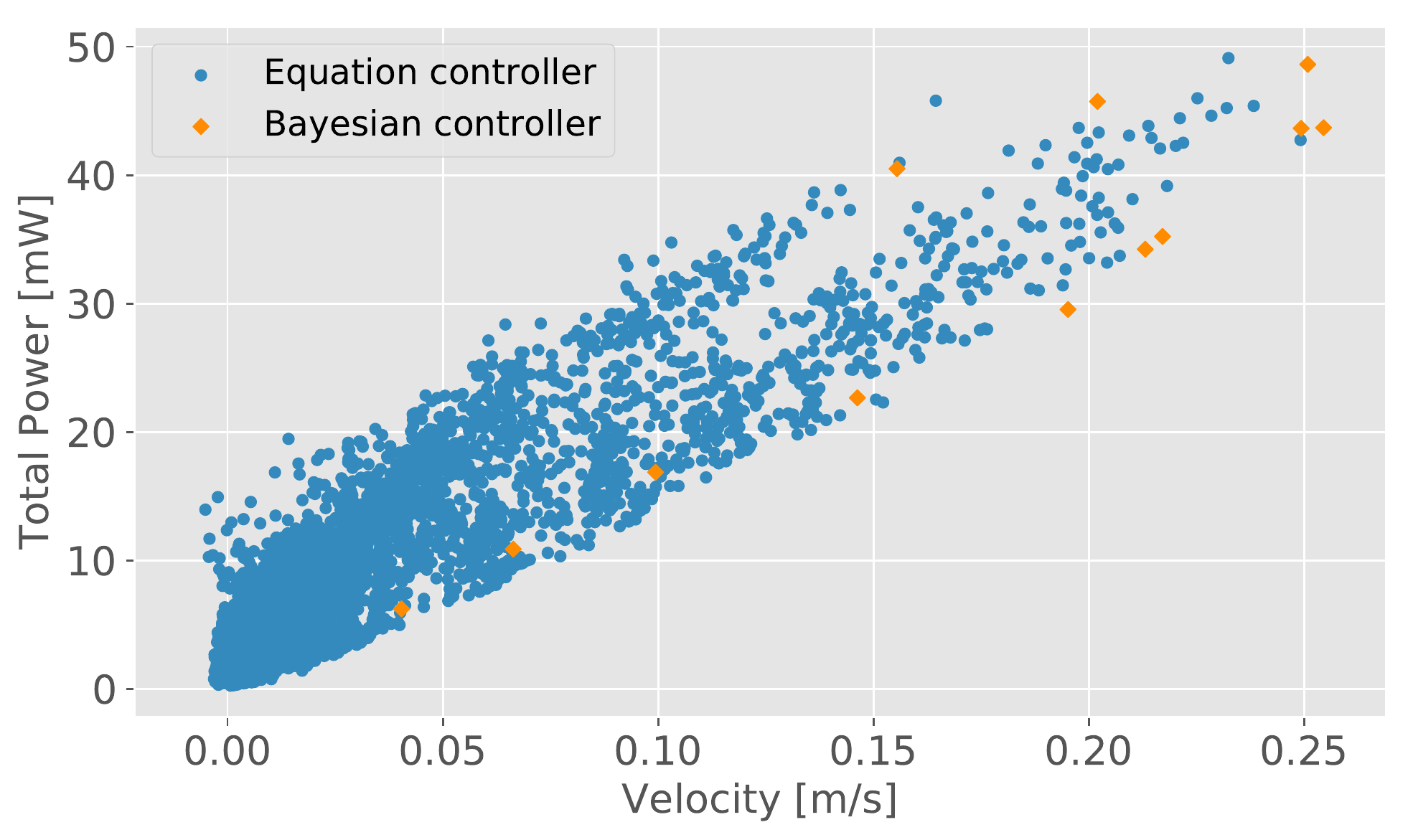}}
	\caption{Baseline examples. This scatter plot shows the results from the grid search algorithm (blue point) and the Bayesian optimization algorithm (orange diamond), respectively. 
		The coordinate of each data point represents the mean velocity and its corresponding power consumption. }
	%\caption{Scatter plot of the parameter grid search of the equation controller. Each parameter set is represented by one data point.}
	\label{fig_grid_search_results_velocity_power_equation}
\end{figure}

The power consumption and the corresponding velocity results from the grid search algorithm and the Bayesian optimization algorithm are shown in Figure~\ref{fig_grid_search_results_velocity_power_equation} as a point cloud of parameter sets using blue dot and orange diamond markers, respectively. 
The lowest points at different velocities in the point cloud have the highest efficiency. 
As we can see, the power requirement grows linearly with the increasing velocity. 
Most results from the Bayesian optimization controller (orange diamond marker) almost match the best-energy-efficiency points from the grid search method, which proves the applicability of the algorithm.
And some points even exhibit better efficiency especially when the desired velocity is higher, which shows the capabilities of the Bayesian optimization algorithm for choosing better parameters for gait generation tasks.
However, there are two outliers with low efficiency, which may be caused by falling into local minimum values during the optimization process.

In short, we demonstrate that the Bayesian optimization algorithm is a quick and effective method to find a proper parameter set for achieving desired velocity, especially when most of the parameter sets in the searching space are distributed in the low velocity area ($0~m/s~\sim~0.1~m/s$).

\section{Proposed RL-Based Controller}
\label{sec_approach}

We begin this section by introducing the key ingredients of our reinforcement learning-based controller.
Then the RL network architecture and the training configuration are introduced as well.

\begin{figure*}[t]
	\centering
	\includegraphics[width=0.8\textwidth]{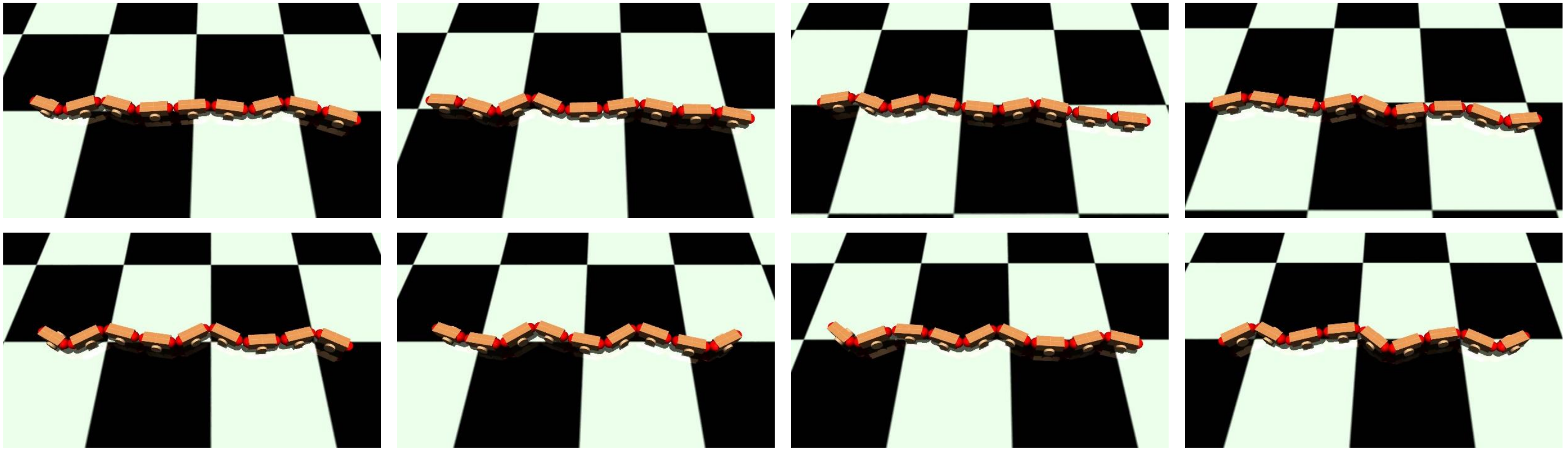}
	\caption{The montages of the snake-like robot model performing two learned gaits at different speeds. 
		The frames are sorted in two rows from left to right and are recorded at intervals of 150 ms (the duration for 3 time steps).
		The first row is captured at a velocity of $0.05~m/s$ and appears to be more similar to the concertina gait.
		The second row is captured at a velocity of $0.25~m/s$ and resembles the slithering gait.}
	\label{fig_montage}
\end{figure*}

\subsection{Reinforcement Learning Setup}
%\textcolor{red}{MISSING Info}
Below we describe the details of the observation space, the action space, and the reward function.

\subsubsection{Observation Space}
The RL agent receives the environmental information via the observation space at each step.
%The observation space is a set of information about the environment, which the agent receives from the environment during each step. 
A proper choice of the observation space parameters is critical in RL, since the agent needs the right set of information to learn the causality of the given rewards based on its actions. 

The observation space $\textbf{o}_{i}^{t}$ is given in Table~\ref{tab:observation_space}. 
The joint position $\phi_j$ and its angular joint velocity $\dot{\phi_j}$ are required to learn the locomotion and represent the proprioceptive awareness of the robot. 
The head link velocity $v_1$ helps to sense its global velocity, which offers better movement awareness. 
%The sense for the target object direction is provided by the relative angle between the head facing direction and the target object $\phi_t$. 
%This is used instead of the head camera to simplify the navigation process, since this experiment focuses on the locomotion. 
In order to learn a power efficient gait, the sense of energy consumption is necessary. 
Therefore, the actuator torque of each joint $\tau_j$ is provided and can be interpreted in combination with $\dot{\phi_j}$ to determine the total power usage. 
%Another important parameter is the actuator torque of each joint $\tau_j$, which is combined with $\dot{\phi_j}$ the basis to give a sense for the used power and enables the learning of an energy efficiency locomotion. 
The specified target velocity $v_t$ is passed to the environment and can be dynamically changed. 
This is required to control the velocity of the robot. 
In summary, an overall 26-DOF observation space is used in this work.

\begin{table}[b]	
	\centering
	\def\arraystretch{1.1}
	\begin{tabular}{l l}
		\toprule
		Symbols & Descriptions\\
		\hline
		\(\phi_{1-8}\) & Relative joint angular positions \\
		\hline
		\\[-1em]
		\(\dot{\phi}_{1-8}\) & Relative joint angular velocity\\
		\hline  
		\(v_1\) & Absolute head module velocity \\
		\hline  
		\(\tau_{1-8}\) & Actuator torque output \\
		%		\hline  
		%		\(\phi_t\) & Relative angle between the \\
		%		& head direction and the target\\
		\hline  
		\(v_t\) & Specified target velocity \\
		\bottomrule 
	\end{tabular}
	\caption{The observation space $\textbf{o}_{i}^{t}$ of the PPO controller.\label{tab:observation_space}}
\end{table}

\subsubsection{Action Space}
The action space $\textbf{a}_{i}^{t}$ of the environment has 8-DOF with finite continuous values in the range of $[-1.5, 1.5]$, which linearly translates to a corresponding joint angle \(\phi\) in range of $[-90^{\circ}, 90^{\circ}]$. 
Each action represents eight actuator angle positions of the servo motors. 
A uniform setup with servo motors is chosen so the environment between the two controllers can be compared.
There has been no significant difference noticed between a servo and a torque setup.

\subsubsection{Reward Function}

The objective of this experiment is to learn a power efficient locomotion for a variety of specified velocities.
%In addition, the locomotion direction is defined by the angle between the head and the target. 
Therefore, the energy consumption and the difference between the actual model velocity and the target velocity are the main criterias to find a successful behavior.
The challenge is to combine and weigh the variables into one numerical reward for each time step. 
%To combine the variables to one reward, it was useful to split the power efficiency and velocity criteria into two normalized reward function components.
Therefore, we first split the power efficiency and velocity criteria into two normalized reward function components.
%The normalization simplifies the weighting of both components. 
%Several attempts have led to this solution, although it can be assumed that better solutions are possible. 
%The solution of the reward function is described in the following paragraphs.

First, a normalized reward is defined to maintain the specified velocities. 
The objective is to reach and maintain the target velocity $v_t$ by comparing it with the head velocity $v_1$.
The following function represents the velocity reward:
\begin{equation}
r_v = (1 - \frac{|v_t - v_1|}{a_1})^{\frac{1}{a_2}}
\end{equation}
The parameter $a_1=0.2$ influences the spread of the reward curve, by defining the x-axis intersections with $x = v_t \pm a_1$. 
$a_2=0.2$ affects the changes of the curves gradient.
If $|v_t - v_1| = 0.0$, the velocity reward $r_v$ has to be the maximum value $1.0$.

%Second, the solution to the power efficiency reward component $r_P$ of the reward function is shown in the following paragraphs. 
Second, the normalized value of the total power usage $\hat{P}$ in \eqref{equ_normal_power} is used to determine the power efficiency reward component $r_P$, 
%$\hat{P}$ has to be optimized to a minimum power usage to achieve the maximum power efficiency. 
%In terms of the actual value range of $\hat{P}$, it has to be considered that the values in the lower quarter are more realistic. 
%This is due to the physical limitations of the snake, where maybe only the joints of the head or the tail are free enough to reach the maximum of $\dot{\phi_j}$. 
%A value range around 0.001 and 0.05 is more likely, based on observations of experiments with power efficient behavior. 
%The power efficiency part of the reward function is represented by
which is represented by
\begin{equation}
\centering
r_P = r_{max} |1 - \hat{P}|^{{b_1}^{-2}}~\text{.}
\label{equ_reward_p}
\end{equation}
Here, $r_{max}$ controls the maximum reward value and $b_1=0.6$ is the slope of the curve. 
The power efficiency is influenced by the desired target velocity. 
Therefore, the normalized value $r_{max}$ represents this influence by limiting the maximum value of $r_P$.

Last, the rewards from the velocity $r_v$ and the power efficiency $r_P$ are combined to form the overall reward $r$:
\begin{equation}
\centering
r = (1 - \frac{|v_t - v|}{a_1})^{\frac{1}{a_2}} |1 - \hat{P}|^{{b_1}^{-2}}
\end{equation}
This equation replaces the $r_{max}$ in \eqref{equ_reward_p} with $r_v$. 
With that, the maximal power efficiency depends on the absolute value of the difference between the desired velocity and the robot velocity. 
%Thus a high dependence is achieved.

\subsection{Network Architecture}
Given the input (observation $\textbf{o}_{i}^{t}$) and the output (action $\textbf{a}_{i}^{t}$), we now elaborate the policy network mapping $\textbf{o}_{i}^{t}$ to $\textbf{a}_{i}^{t}$.
We design a fully connected 2-hidden-layer neural network as a non-linear function approximator to the policy $\pi_{\theta}$.
The input layer has the same dimension as the action space $\textbf{o}_{i}^{t}$.
Both hidden layers have $200$ ReLU units and the final layer outputs the joint position commands for the robot.
In order to train the network, the PPO algorithm adapted from~\cite{schulman2017proximal} is used for training it.

\subsection{Training Configuration}
To enable the learning of different velocities, the parameter $v_t$ is changed by iterating over $0.05$, $0.1$, $0.15$, $0.20$, and $0.25$ for each episode while training. 
Meanwhile, to simplify the beginning of the learning process the first $100$ episodes are trained with a fixed target velocity of $0.1~m/s$.

We train our policy network on a computer with an i7-7700 CPU and a Nvidia GTX 1080 GPU. 
Based on the learning curve a total of 3 million time steps (about $1400$ updates) are used for training. 
With the environment settings of $50~ms$ per time step, the training takes about $42$ hours in total simulation time and $2$ hours in wall clock time for the policy to converge to achieve a robust performance.

\section{Results and Comparisons}
\label{sec_experiment}
In this section, we first describe the performance of the gaits generated by the RL controller. 
Then, we compare our gaits to the scripted slithering gaits in terms of energy efficiency.  

%\subsection{Training Configuration}
%
%We train our policy network on a computer with an i7-7700 CPU and a Nvidia GTX 1080 GPU. 
%Based on the learning curve a total of 3 million time steps (about $1400$ updates) are used for training. 
%With the environment settings of $50~ms$ per time step, the training takes about $42$ hours in total simulation time and $2$ hours wall clock time for the policy to converge to a robust performance. 
%%MuJoCo, thanks to its fast and accurate simulation, was able to speed up the simulation time to a duration of about $2.0~h$ elapsed real time on a personal computer with a $4.0$ GHz CPU.

\subsection{Results}
In this study, $45$ target velocities in the range of $[0.025, 0.25]$ with a step interval of $0.005$ are used for the evaluation of performance. 
Simulation results demonstrate that the PPO controller has succeeded in learning a gait from scratch without knowing any prior locomotion skills.
Two exemplary gait patterns are shown in Figure~\ref{fig_montage}.
Surprisingly, we find that the PPO controller can adapt its gait pattern according to its desired speed like a real snake.
The montages in the first row present a concertina-like gait at a low speed of $0.05~m/s$ and the second row resembles a slithering-like gait at a higher speed of $0.25~m/s$.
In nature, snakes usually take the concertina locomotion at a low speed and switch to the slithering locomotion at a fast speed.
In the second row, we also observe that the snake-like robot executes waves from the head to tail through a movement pattern of lateral undulation.
This learned undulation wave is even smoother than the parameterized \textit{serpentine} curve and drives the robot to move with better power efficiency.

\begin{figure}[t]
	\centerline{
		\includegraphics[width=0.45\textwidth]{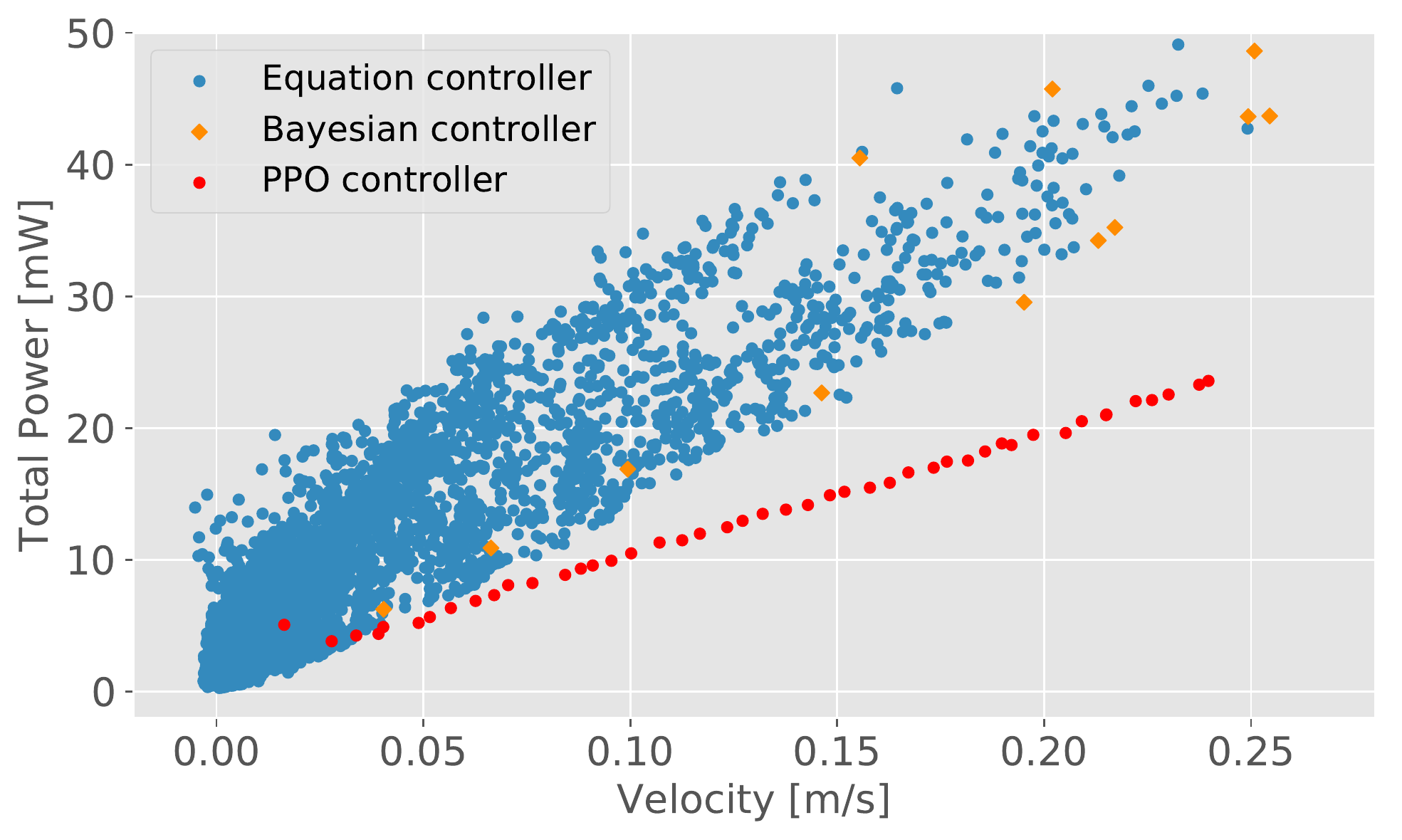}}
	\caption{This scatter plot directly shows the energy consumption results of the controllers at a range of velocities. The blue point, orange diamond, and the red point stand for the data from the gait equation controller, Bayesian optimization controller, and the PPO controller, respectively.}
	\label{fig:velocity_power_comparison}
\end{figure}

The power consumption results of the learned gait are marked with red points in Figure~\ref{fig:velocity_power_comparison}.
The data depicts a linear relationship between the travel velocity and the power consumption, which is in line with the physical law $Power = Force \times Speed$.
There is only one point with higher power consumption when the velocity is around $0.02~m/s$. 
Importantly, this also reveals the adaptability of the learning approach for generating gaits in a range of velocities.
It can also be observed that the mean velocities do not exactly align with the specified interval of $0.005$, especially at higher target velocities.
The reason for this is the difficulty to achieve the exact ratio between holding the right velocity and performing the corresponding power-efficient locomotion.

\subsection{Comparison}

We first have a look at the energy metric based the averaged power per velocity. 
After putting the power efficiency data of the \textit{gait equation} controller (See Figure~\ref{fig_grid_search_results_velocity_power_equation}) and the PPO controller together, we can clearly conclude that the PPO controller has a much better power efficiency at a range of velocities (See Figure~\ref{fig:velocity_power_comparison}).
As the velocity grows, the advantage of the PPO controller for saving energy is even more obvious.
Taking the velocity of $0.15~m/s$ as an example, the PPO controller can save $35\%$ $\sim$ $65\%$ energy consumption depending on the parameter sets chosen by the \textit{gait equation} controller.

\begin{figure}[t]
	\centering
	\includegraphics[width=0.45\textwidth]{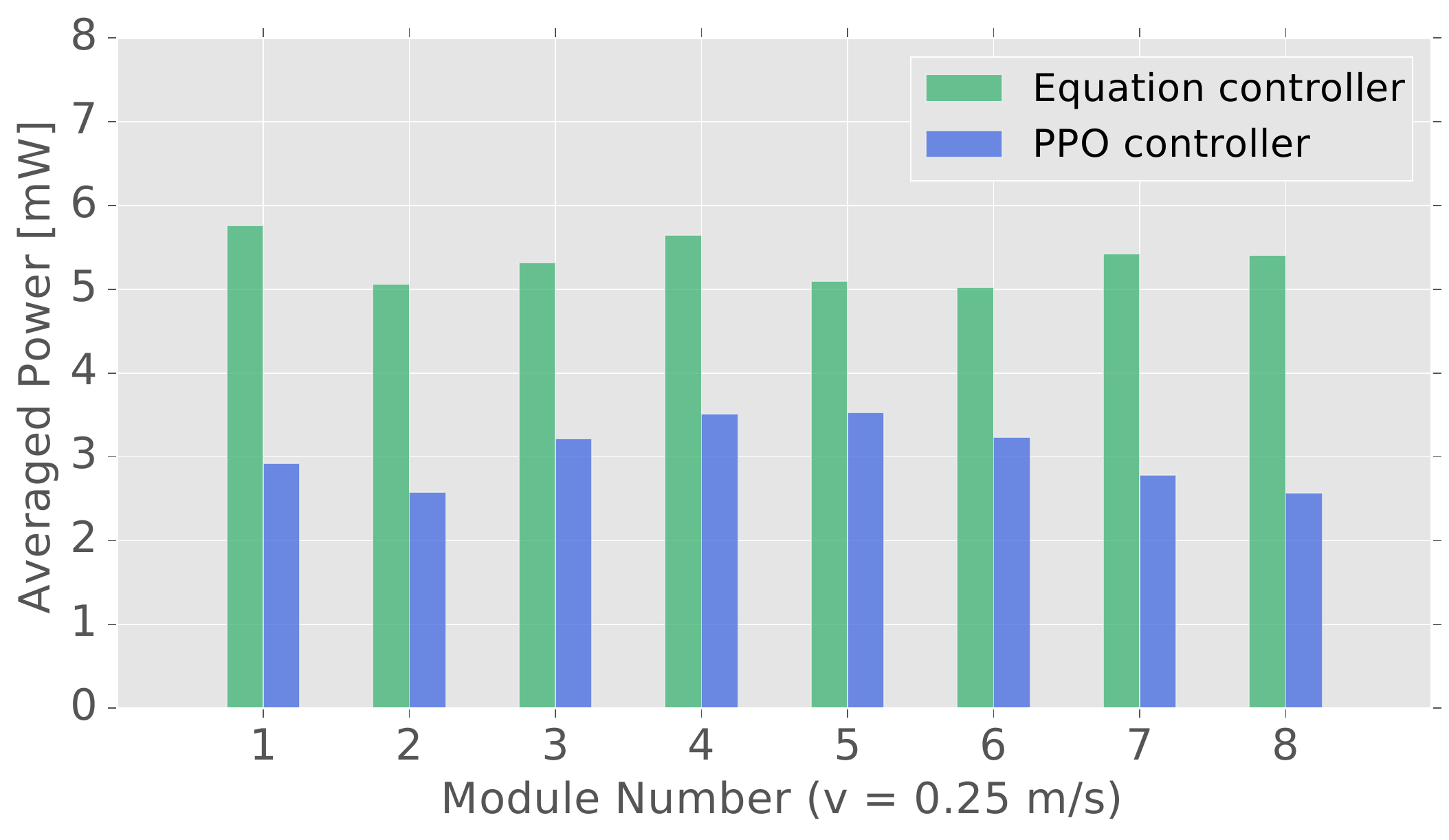}
	\caption{The power profile bar of the slithering gait at the velocity of $0.25~m/s$ for the equation controller and the PPO controller.}
	\label{fig_joint_power}
\end{figure}

We then discuss the power profile for each module of the robot. 
The power profiles of the slithering gait at the velocity of $0.25~m/s$ generated from the parameterized equation controller and the PPO controller are shown in Figure~\ref{fig_joint_power}.
It should noted that we only present the results of the parameterized equation controller with the best energy efficiency at this speed.
For the parameterized equation controller, the first joint consumes the most power since it rapidly adjust its head to aim at the front direction. 
Other module joints use similar power around $5 mW$ to $5.7 mW$.
For the PPO controller, the head joint still expends more energy than the average power consumption.
Besides, the joints close to the body center consume more power than the far-ends of the body.
This is because the body trunk modules exert more strength to twist its body to generate the locomotion, which shows that the PPO controller generates more natural slithering gait compared to the parameterized slithering gait.

These superiorities can be elaborated from two aspects.
On one hand, the traditional \textit{gait equation} controller is based on the kinematics and describes the gait movement with no influence of physical forces such as friction or damping.
Meanwhile, it only extracts several critical parameters to represent the gait, which is in fact a highly dimensional interaction with the environment.
Although this parameterized description simplifies the control task, it inevitably brings difficulties for designing more sophisticated gaits despite of using optimization technologies, especially when the parameter space will grow exponentially with the increasing joint numbers.
On the other hand, the RL method shows its effectiveness in the ability for solving this kind of complex control problem, since it is trained directly in the dynamic environment. 
It is able to generate an undulation gait that not only imitates real snakes, since natural evolution is not a perfect process but a compromise result.
Therefore, it can explore its limitations and keep improving its behavior under its hardware constraints.
This is a remarkable achievement because the algorithm has to handle a high degree of freedom without prior information about the environment or any locomotion behavior. 
In short, the PPO controller is able to overall outperform the equation controller in terms of power efficiency at a range of velocities.

%Since the equation controller is based on a kinematic model, the controllers can only be compared to a limited extent. 
%As a kinematic model, it describes the gait movement without any influence of physical forces. 
%This is why such a model is not optimal applicable for dynamical simulation environments.
%However, in the case of the used snake-like robot model, it is less critical to use a kinematic model controller. 
%For instance, the usage of free wheels limits the fiction forces influence to the ground and keeps the snake-like robot model balanced. 
%This would not be the case with a legged robot, as it requires a dynamical controller for its balance. 
%Furthermore, the simulation environment has been customized to be able to use the equation controller. 
%Under concern of these restrictions, the PPO controller is limited comparable to the equation controller. 
%Nevertheless, the PPO controller was able to overall outperform the equation controller in terms of power efficiency.

\section{Conclusion}
\label{sec_conclusion}

Designing power-efficient gaits for snake-like robots remains a challenging task, since they come with redundant degrees of freedom and have complicated interactions with the environment. 
In this paper, we present a novel gait design method based on reinforcement learning.
The learned gait has shown to have much better energy efficiencies at different travel velocities compared to the current kinematic-based method even after choosing those parameters using the grid search or Bayesian optimization algorithm.
Our work contributes and serves as an exploration for designing sophisticated moving patterns for snake-like robots.
Our future work will aim at designing gaits based on reinforcement learning for those kinds of snake-like robots without passive wheels.

\begin{spacing}{0.95}
\bibliographystyle{named}
\bibliography{ijcai19}
\end{spacing}

\end{document}